\documentclass{egpubl}
\usepackage{egsr18-e}

\WsPaper         
 \electronicVersion 

\ifpdf \usepackage[pdftex]{graphicx} \pdfcompresslevel=9
\else \usepackage[dvips]{graphicx} \fi

\PrintedOrElectronic

\usepackage{t1enc,dfadobe}

\usepackage{egweblnk}
\usepackage{cite}

\usepackage{t1enc,dfadobe}

\usepackage{booktabs} 
\usepackage{egweblnk}
\usepackage{cite}

\usepackage{amsmath}
\usepackage{stmaryrd}

\newcommand{\argmin}{\mathop{\mathrm{argmin}}}

\usepackage{makecell}
\usepackage{pifont}
\newcommand{\cmark}{\ding{51}}%
\newcommand{\xmark}{\ding{55}}%

\title[Deep Hybrid Real and Synthetic Training for Intrinsic Decomposition]%
      {Deep Hybrid Real and Synthetic Training \\~for  Intrinsic Decomposition}

\author[S. Bi, N. Khademi Kalantari, R. Ramamoorthi]
{
   \parbox{\textwidth}{\centering 
       Sai Bi, 
       Nima Khademi Kalantari,
       Ravi Ramamoorthi    
   }
   \\
    {\parbox{\textwidth}{\centering 
       University of California, San Diego
        }
    }
}

\begin{document}

\teaser{
\centering
\includegraphics[width=\linewidth]{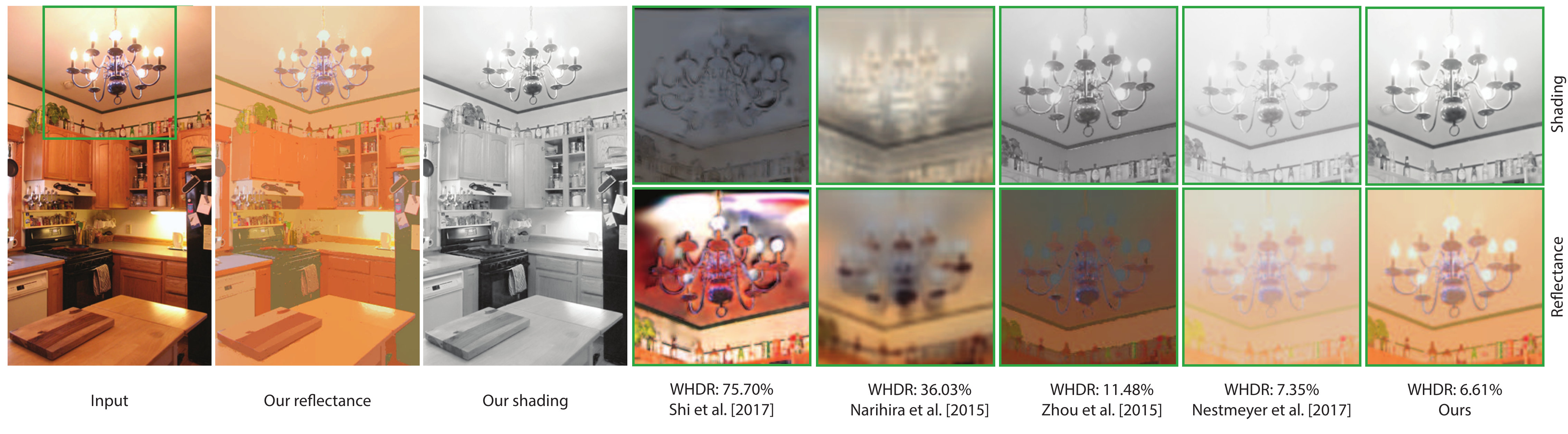}
\caption{
Our method decomposes an input image into intrinsic
reflectance and shading layers. We propose a novel hybrid learning approach to
train a convolutional neural network (CNN) on both synthetic and real images.
To utilize the real images, we rely on the fact that two images of a scene
captured with different lightings have the same reflectance, and thus, enforce
that the network produces consistent results for the real image pairs during
training. Here, we compare our approach against several state-of-the-art
methods on a test image from the IIW dataset~\cite{bell14}. Note that our
system is not trained on this dataset. The deep learning (DL) approaches by Narihira et
al.~\shortcite{narihira15} and Shi et al.~\shortcite{shi17} train a CNN only on
synthetic data, and thus, are not able to produce reasonable results on real
images because of the large gap in data distributions. Although the learning-based
(non-DL) methods by Zhou et
al.~\shortcite{zhou15} and Nestmeyer and Gehler~\shortcite{nest17} specifically
train their system on the IIW dataset, they are not able to handle the large
shading variations in this scene. Note the brightness inconsistency of the
ceiling in their reflectance insets. In comparison, our hybrid training
approach is able to generate better results both visually and numerically, in
terms of WHDR scores~\cite{bell14} (lower is better). Note that the WHDR scores
are calculated on the full images. See supplementary materials for the full
images.  We also compare against non-learning approaches in Table~\ref{fig:whdr} and Fig.~\ref{fig:iiw-comp}.}
\label{fig:Teaser}
}

\maketitle

\begin{abstract}
Intrinsic image decomposition is the process of separating the reflectance and
shading layers of an image, which is a challenging and underdetermined problem.
In this paper, we propose to systematically address this problem using a deep
convolutional neural network (CNN). Although deep learning (DL) has been recently
used to handle this application, the current DL methods train the
network only on synthetic images as obtaining ground truth reflectance and
shading for real images is difficult. Therefore, these methods fail to produce
reasonable results on real images and often perform worse than the non-DL techniques. 
We overcome this limitation by proposing a
novel hybrid approach to train our network on both synthetic and real images.
Specifically, in addition to directly supervising the network using synthetic
images, we train the network by enforcing it to produce the same reflectance
for a pair of images of the same real-world scene with different illuminations.
Furthermore, we improve the results by incorporating a bilateral solver layer into our
system during both training and test stages. Experimental results show that our
approach produces better results than the state-of-the-art DL and non-DL methods on various
synthetic and real datasets both visually and numerically. 


\end{abstract}  
\section{Introduction}
\label{sec:Introduction}
The visual appearance of objects in images is determined by the interactions
between illumination and physical properties of the objects such as their
materials. Separating an image into  reflectance and shading layers is called
intrinsic decomposition and has a wide range of computer graphics/vision
applications~\cite{Bonneel17}, such as surface retexturing, scene relighting, and material
recognition. The intrinsic image model states that the input image is the
product of the reflectance and shading images, and thus, the problem of
inferring these two images from the input image is heavily underdetermined.

\begin{figure}[t]
    \centering
    \includegraphics[width=\linewidth]{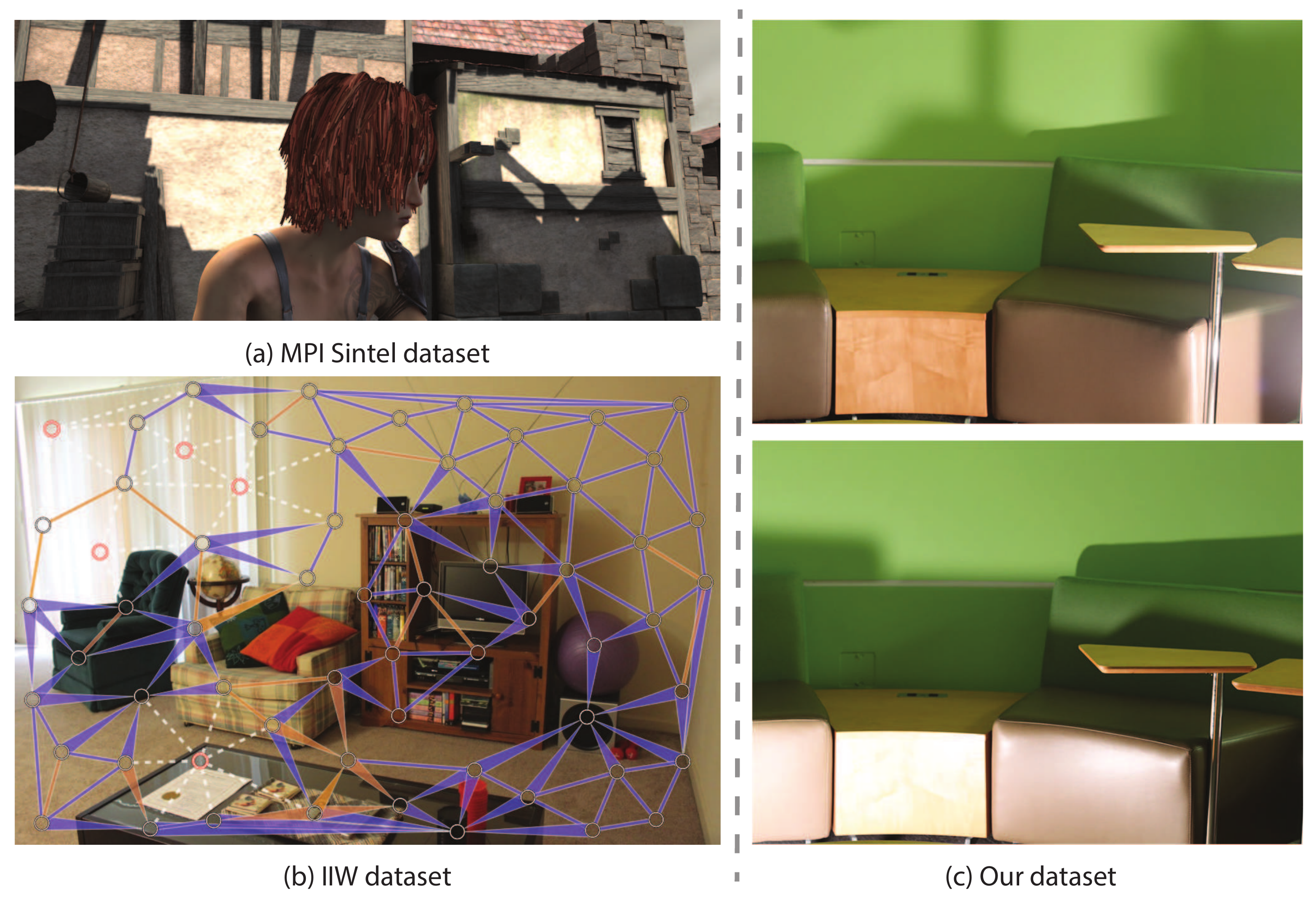}
    \caption{\label{fig:data}
    Several existing learning-based approaches train their system only on
    synthetic images (a), and thus, produce sub-optimal results on real images.
    Others use the pairwise relationships between reflectance values over pixel
    pairs from the IIW dataset (b), which is not sufficient to train a reliable
    model. On the other hand, we train our system on both synthetic images from the
    Sintel dataset and pairs of real world images of the same scene with different
    illuminations (c).  }
\end{figure}

A category of successful methods addresses this problem in a data-driven way. A
few approaches~\cite{narihira15,shi17} propose to train a model only on
synthetic images, like the one shown in Fig.~\ref{fig:data}-a, as obtaining
ground truth reflectance and shading for real images is difficult. However,
since the data distribution of synthetic and real images is different, these
methods produce sub-optimal results on real images, as shown in
Fig.~\ref{fig:Teaser}. Although several approaches~\cite{zhou15,nest17} train
their system on real images from the IIW dataset~\cite{bell14}, this dataset
only contains relative comparison over reflectance values at pixel pairs (see
Fig.~\ref{fig:data}-b), which is not sufficient for training a reliable model.
Therefore, these methods often are not able to fully separate shading from the
reflectance, as shown in Fig.~\ref{fig:Teaser}.

To address these problems, we propose a hybrid approach to train a
convolutional neural network (CNN) on both synthetic and real images. Our main
observation is that a pair of images of the same scene with different
illumination (see Fig.~\ref{fig:data}-c) have the same reflectance. We use this
property to train our network on real images by enforcing the CNN to produce
the same reflectance for the pair of images. In our system, the synthetic data
constrains the network to produce meaningful outputs, while the real data tunes
the network to produce high-quality results on real-world images. To improve
the spatial coherency of the results, we propose to incorporate a bilateral
solver layer with our network during both the training and test stages. We extensively
evaluate our approach on a variety of synthetic and real datasets and show
significant improvement over state-of-the-art methods, both visually
(Figs.~\ref{fig:Teaser}~and~\ref{fig:iiw-comp}) and numerically
(Tables~\ref{fig:whdr}~and~\ref{fig:syn-score}). 
In summary, we make the following contributions:

\begin{enumerate}
    \item 
        We propose a hybrid approach to train a CNN on both synthetic and real
        images (Sec.~\ref{ssec:Training}) to address the shortcomings of previous
        approaches that train their models only on synthetic data or on real images
    with limited pairwise comparisons.  
    \item We incorporate a bilateral solver layer into
        our network and train it in an end-to-end fashion to suppress the potential
        noise and improve the spatial coherency of the results
        (Sec.~\ref{ssec:NetArch}).
\end{enumerate}

\section{Related Work}
\label{sec:RelatedWork}
Barrow and Tenenbaum~\shortcite{barron78} introduced the concept of intrinsic
decomposition and showed that it is desirable and possible to describe a scene
in terms of its intrinsic characteristics. Since then significant research has
been done and many powerful approaches have been developed. Several methods
have proposed more complex imaging models and decompose the image into
specular~\cite{shi17} or ambient occlusion~\cite{hauagge16, Carlo17} layers in addition
to the typical reflectance and shading layers. Here, we consider a standard
intrinsic image model, which covers the majority of real-world cases, and
decompose the image into reflectance and shading layers. Moreover, several
methods utilize additional information like depth~\cite{chen13}, user
interactions~\cite{Bousseau09}, or use a
collection of images~\cite{laffont12} to facilitate intrinsic decomposition.
For brevity, we only focus on approaches that perform intrinsic decomposition
on a single input image by categorizing them into two general classes of
physical prior based and data-driven approaches.

{\em Physical Prior Based Approaches --} The approaches in this category use
a physical prior in their optimization system to address this underdetermined
problem. Retinex theory~\cite{land71, Horn74}, the most commonly-used prior, states
that color changes caused by reflectance are generally abrupt, while those from
shading variations are continuous and smooth. Based on this observation, Tappen
et al.~\shortcite{tappen05} train a classifier to determine if the image
derivative at each pixel is because of the shading or reflectance changes. Shen
et al.~\shortcite{shen08} and Zhao et al.~\shortcite{zhao12} observe that pixels with the same local texture
configurations generally have the same reflectance values, and utilize this
observation as a non-local constraint to reduce the number of unknowns when
solving for reflectance. 
A couple of approaches~\cite{pfister14, meka16}
separate image derivatives into smooth shading variations and sparse
reflectance changes by encoding the priors in a hybrid $\ell_2$-$\ell_p$
objective function. Zoran et al.~\shortcite{zoran15} train a network on the IIW
dataset to predict the ordinal relationship of reflectance values between a
pair of pixels and combine such constraints with Retinex theory to formulate a
quadratic energy function.

Reflectance sparseness is another commonly-used prior, which states that there
are a small number of different reflectance values in natural
images~\cite{omer04}. Barron et al.~\shortcite{barron15} encode this prior 
in their optimization framework by minimizing the global entropy of log-reflectance.
Garces et al.~\shortcite{Garces12}, Bell et al.~\shortcite{bell14}
and Bi et al.~\shortcite{bi15} sample a sparse set of reflectance values from
the image and assign each pixel to a specific value with methods such as
k-means clustering and conditional random
field (CRF). Zhou et al.~\shortcite{zhou15} further improve the CRF framework
by replacing the handcrafted pairwise terms used to evaluate the similarity of
pixel reflectance with the predictions of a neural network trained on the IIW
dataset.

A common problem with the approaches mentioned above is that the priors are not
general enough to cover a wide range of real world complex scenes. For example,
shading smoothness does not always hold in practice due to the existence of
depth discontinuities, occlusions, as well as abrupt surface normal changes. In
this case, approaches based on these priors will inevitably generate incorrect
decompositions. In contrast, our approach is fully data-driven and does not
have these problems. 

{\em Data-Driven Approaches --} Recently some approaches have tried to directly
estimate the reflectance and shading by training a CNN. The major challenge is
that obtaining ground truth reflectance and shading for real-world scenes is
difficult. To overcome this limitation, approaches~\cite{narihira15,shi17} have
been proposed to train a CNN on synthetic datasets, such as MPI
Sintel~\cite{butler12} and ShapeNet~\cite{chang15}. These approaches perform
poorly on real images because of the large gap between the distribution of
synthetic and real-world data. Nestmeyer and Gehler~\shortcite{nest17} train a
network utilizing only pairwise ordinal relationships on reflectance values in
the IIW dataset, which only contains sparse annotations over a small number of
pixels. However, these pairwise comparisons do not provide sufficient
supervision for the network to produce high-quality results in challenging
cases (Figs.~\ref{fig:Teaser}~and~\ref{fig:iiw-comp}). We avoid these problems
by proposing a novel approach to utilize real-world images for supervising the
network and perform a hybrid training on both real and synthetic datasets.

\section{Algorithm}
\label{sec:Algorithm}

The goal of our approach is to decompose an input image, $I$, into its
corresponding reflectance, $R$, and shading, $S$, images. In our system, we
consider the input image to be the product of the reflectance and shading
layers, i.e., $I = R\cdot S$. For simplicity we assume the scenes to have 
achromatic lighting (i.e. $S$ is a single channel image) similar to the 
majority of existing techniques. However, extension to colored
lighting is straightforward. Inspired by the success of deep learning in a
variety of applications, we propose to model this complex process using a CNN.
The key to success of every learning system lies in effective training and
appropriate network architecture, which we discuss in
Secs.~\ref{ssec:Training}~and~\ref{ssec:NetArch}, respectively.

\subsection{Training}
\label{ssec:Training}

\begin{figure}[t]
  \centering
  \includegraphics[width=\linewidth]{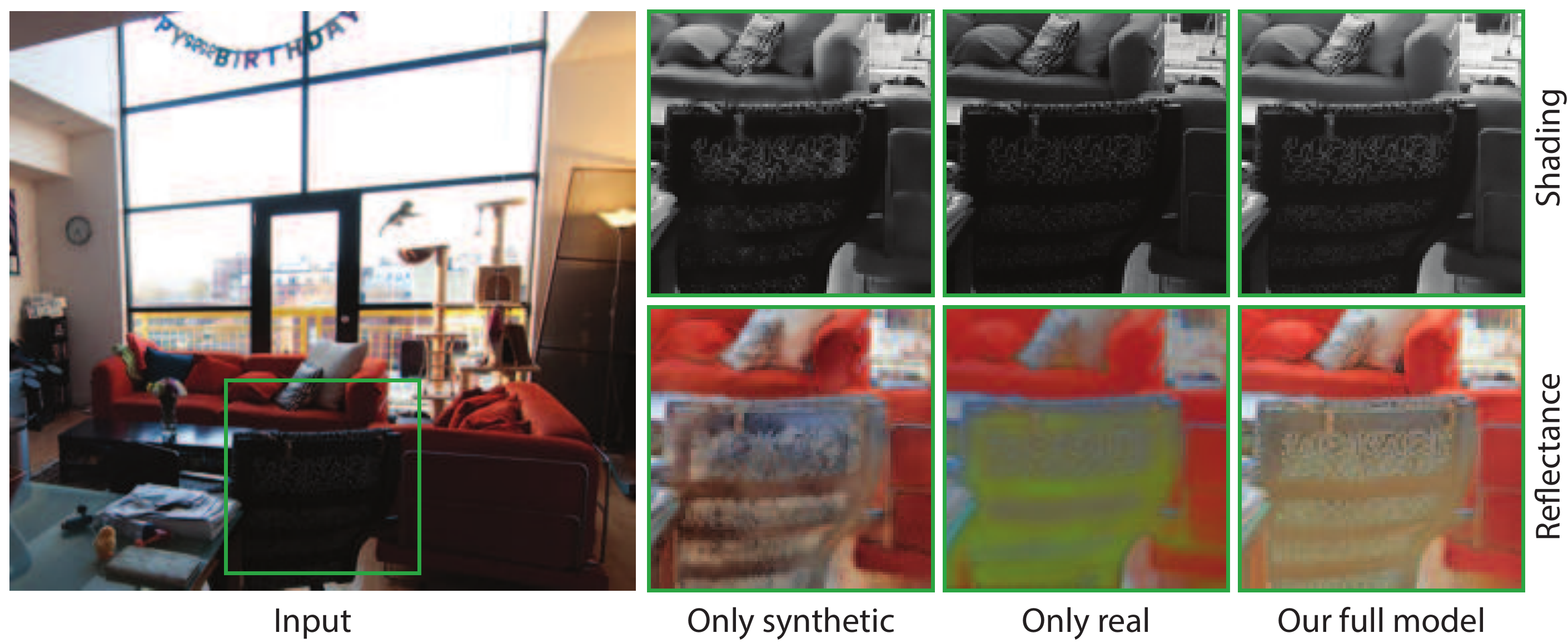}
  \caption{\label{fig:intra-2} 
  We analyze different terms in Eq.~\ref{eq:loss} by training the system on
  each term and comparing it to our full approach. As shown in the insets, the
  synthetic and real losses alone are not sufficient to produce high-quality
  results. Our approach minimizes both terms, and thus, produces reflectance
  and shading images with higher quality.  }
\end{figure}

As discussed, obtaining ground truth reflectance and shading images for
real-world scenes is difficult. To overcome this limitation, we make a key
observation that two images of the same scene under different illuminations
have the same reflectance. Based on this observation, we propose a novel
approach to provide a weak supervision for the network by enforcing it to
produce the same reflectance for pairs of real images. Although necessary, this
weak supervision is not sufficient for training a reliable model, as shown in
Fig.~\ref{fig:intra-2}. Our main contribution is to propose a hybrid training
on both synthetic and real images by minimizing the following loss:

\begin{equation}
    E = E_{\text{syn}} + \omega E_{\text{real}},
  \label{eq:loss}
\end{equation}

\noindent where $E_{\text{syn}}$ and $E_{\text{real}}$ are defined in
Eqs.~\ref{eq:synthetic}~and~\ref{eq:real}, respectively, and $\omega$ defines
the weight of the real loss. We set $\omega$ to 0.5 to keep the real and
synthetic losses within a reasonable range and avoid instability in training.
Intuitively, training on synthetic images constrains the network to produce
meaningful outputs, while the weak supervision on real images adapts the
network to real-world scenes. Note that, as shown in Fig.~\ref{fig:intra-2}, a
network trained only on synthetic images does not generalize well to real
images and both terms in Eq.~\ref{eq:loss} are necessary to produce
high-quality results. Next, we discuss our synthetic ($E_{\text{syn}}$) and
real ($E_{\text{real}}$) losses, as well as the training details and dataset.

{\em Synthetic Images --} Since the ground truth reflectance, $\hat{R}$, and
shading, $\hat{S}$, images for a synthetic dataset are available, we can
provide a direct supervision for the network by minimizing the mean squared
error (MSE) between the ground truth and the estimated reflectance and shading
layers. However, there exists an inherent scale ambiguity in intrinsic
decomposition, i.e., if $R$ and $S$ are the true reflectance and shading
images, $\alpha R$ and $S/\alpha$ are also valid. Therefore, we use the scale
invariant MSE ($E_{\text{si}}$)~\cite{eigen14} to measure the quality of our
decomposition. Our synthetic loss is defined as follows:

\begin{equation}
  E_{\text{syn}} = E_{\text{si}}(R, \hat{R}) + E_{\text{si}}(S, \hat{S}) + E_{\text{si}}(R\cdot S, I),
  \label{eq:synthetic}
\end{equation}

\noindent where the scale invariant loss is defined as:

\begin{equation}
\label{eq:simse}
E_{\text{si}}(\hat{M}, M) = \frac{1}{N}||\hat{M} - \alpha M||_2, \enskip \text{where} \enskip \alpha = \argmin_{\alpha} ||\hat{M} - \alpha M||_2.
\end{equation}

\noindent Here, $N$ is the number of elements in $M$, and $\alpha$ is a scalar
with an analytical solution $\alpha = (\sum{\hat{M} \cdot M})/(\sum{M \cdot
M})$.

The first and second terms of Eq.~\ref{eq:synthetic} are necessary to enforce
that the network produces similar results to the ground truth. Moreover, since our
network estimates the reflectance and shading independently, we use the third
term to enforce that the product of them is the same as the original input image.
Note that since our estimated reflectance and shading could potentially have
arbitrary scales, we also use a scale invariant MSE to compare their product
with the input image.


{\em Real Image Pairs --} Since the ground truth reflectance and shading
images are not available in this case, we propose to provide a weak supervision
for training the network by enforcing it to produce the same reflectance for
real image pairs. Specifically, given a pair of real images taken of the same
scene under different illuminations, $I_1$ and $I_2$, we use our network to
estimate their corresponding reflectance ($R_1$ and $R_2$) and shading ($S_1$
and $S_2$) images. We then optimize the following loss function to train our
network:

\begin{equation}
  E_{\text{real}} = E_{\text{si}}(R_1, R_2) + E_{\text{si}}(R_1 \cdot S_2, I_2) + E_{\text{si}}(R_2 \cdot S_1, I_1).
  \label{eq:real}
\end{equation}

\noindent Here, the first term enforces that the network produces the same
reflectance layers for the pair of real images. Note that we use the scale
invariant MSE to reduce the complexity of training and avoid enforcing the
network to unnecessarily resolve the scale ambiguity. This term only provides
feedback for estimating reflectance on the real images, but we also need to
supervise the network for estimating the shading.

We do so by introducing the second and third terms in Eq.~\ref{eq:real}. The
basic idea behind these terms is that the product of the estimated reflectance
and shading should be equal to the corresponding input image, e.g., $R_1 \cdot
S_1 = I_1$. Therefore, we can provide supervision by minimizing the error
between the product of decomposed layers and the input image. Here, we swap the
two reflectance images to further enforce the similarity of the estimated
reflectance images. Note that, as in the synthetic loss, we use the scale
invariant MSE as the estimated reflectance and shading could have arbitrary
scales.

We note that existing methods~\cite{weiss01, laffont12} that use multiple
images for intrinsic decomposition are fundamentally different from ours. These
approaches always require multiple images to perform intrinsic decomposition.
In contrast, we only use the image pairs during training and at the test time,
our trained network works on a single image, and thus, is more practical.

{\em Training Details and Dataset --} Training a network from scratch on
both synthetic and real images by minimizing Eq.~\ref{eq:loss} is challenging.
Therefore, we propose to perform the training in two stages. First, we train
the network only on synthetic images by minimizing the synthetic loss in
Eq.~\ref{eq:synthetic}. After convergence, the network is able to produce
meaningful reflectance and shading images. This stage basically provides a good
initialization for the second stage in which we train the network on both
synthetic and real image pairs by minimizing our full loss in
Eq.~\ref{eq:loss}. The second stage refines the network to generate more
accurate reflectance and shading layers on real images, while preserving its
accuracy on synthetic images.

\begin{figure}[t]
	\centering
	\includegraphics[width=0.95\linewidth]{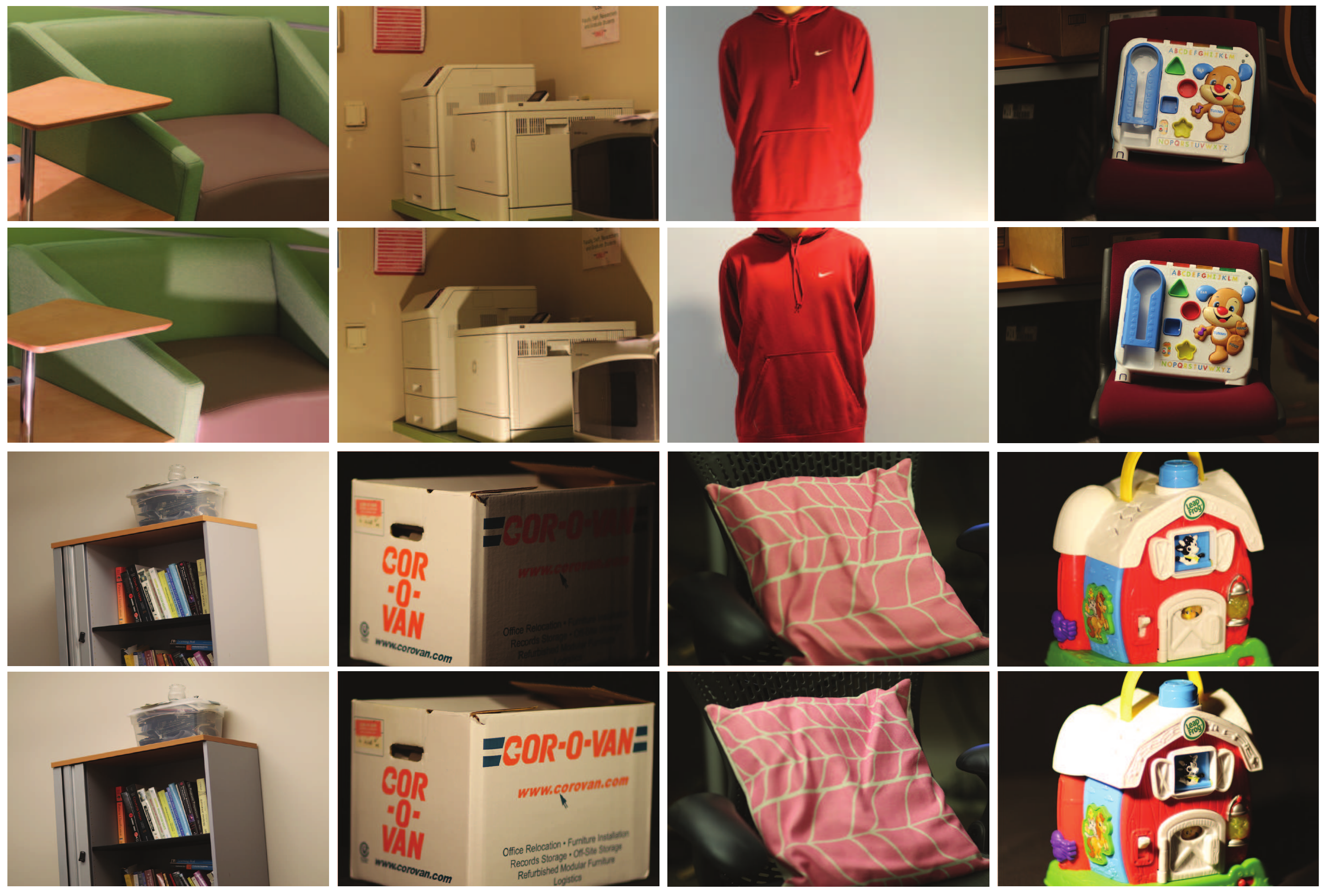}
    \caption{\label{fig:our-dataset} We show real image pairs for a few scenes
from our training set. The images for each scene are obtained by changing the
illumination.  }
\end{figure}

For the synthetic dataset, we use the training images from the MPI Sintel,
containing 440 images of size $1024 \times 436$. For the real images, we used a
tripod mounted camera to take multiple images of a static scene by changing the
illumination of the scene. In order to change the illumination, we use between
one to four light sources and randomly moved them around the scene. We use this
approach to take between 3 to 8 images with varying illumination of 40 scenes.
Note that during every iteration of training we randomly choose only two images
of each scene to minimize Eq.~\ref{eq:loss}. Figure~\ref{fig:our-dataset} shows
image pairs of a few scenes from our training dataset. In
Sec.~\ref{sec:Results}, we demonstrate the performance of our approach on
several real and synthetic scenes, none of them included in the training set.

To increase the efficiency of training, we randomly crop patches of size $256
\times 256$ from the input images. In stage one of the training, we use
mini-batches of size 4, while we perform hybrid training on mini-batches of
size 8 (4 synthetic and 4 real). We implemented our network using the PyTorch
framework, and used ADAM solver~\cite{kingma14} to optimize the network with
the default parameters ($\beta_1 = 0.9, \beta_2 = 0.999$), and a learning rate
of $0.0002$. The training in the first and second stages converged after
roughly 300K and 1.2M iterations, respectively. 
We used an Intel Core i7 with a
GTX 1080Ti GPU to train our network for roughly two days.

\subsection{Network Architecture}
\label{ssec:NetArch}

\begin{figure}[t]
	\centering
	\includegraphics[width=\linewidth]{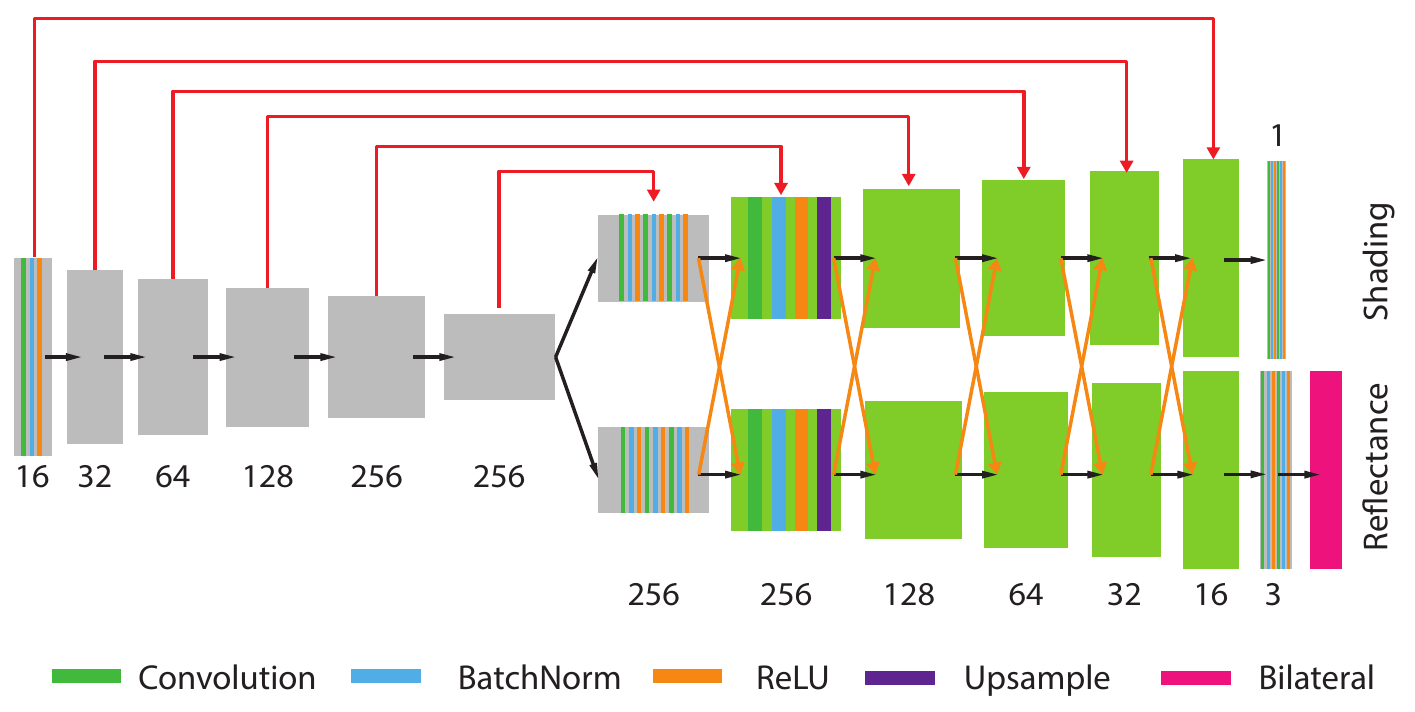}
	\caption{\label{fig:network}
 We use an encoder-decoder network with skip links to train our model.
 Each basic block in the encoder network consists of three types of layers,
 including convolutional layer, batch normalization layer, and 
 Rectified Linear Unit (ReLU) layer. 
 The basic block in the decoder network
 includes an extra bilinear upsampling layer to increase the spatial resolution of
 output. 
}
\end{figure}

We use a CNN with encoder-decoder architecture to model the process, as shown
in Fig.~\ref{fig:network}. Our network contains one encoder, but two decoders
for estimating the reflectance and shading layers. The input to our network is
a single color image with three channels, while the output reflectance and
shading have three and one channel, respectively. Note that our estimated
shading has a single channel since we assume that the lighting is achromatic,
which is a commonly-used assumption~\cite{gehler11}. Since the Sintel images
have colored shading, we first convert them to grayscale and then use them as
ground truth shading for training our network.

\begin{figure}[t]
  \centering
  \includegraphics[width=\linewidth]{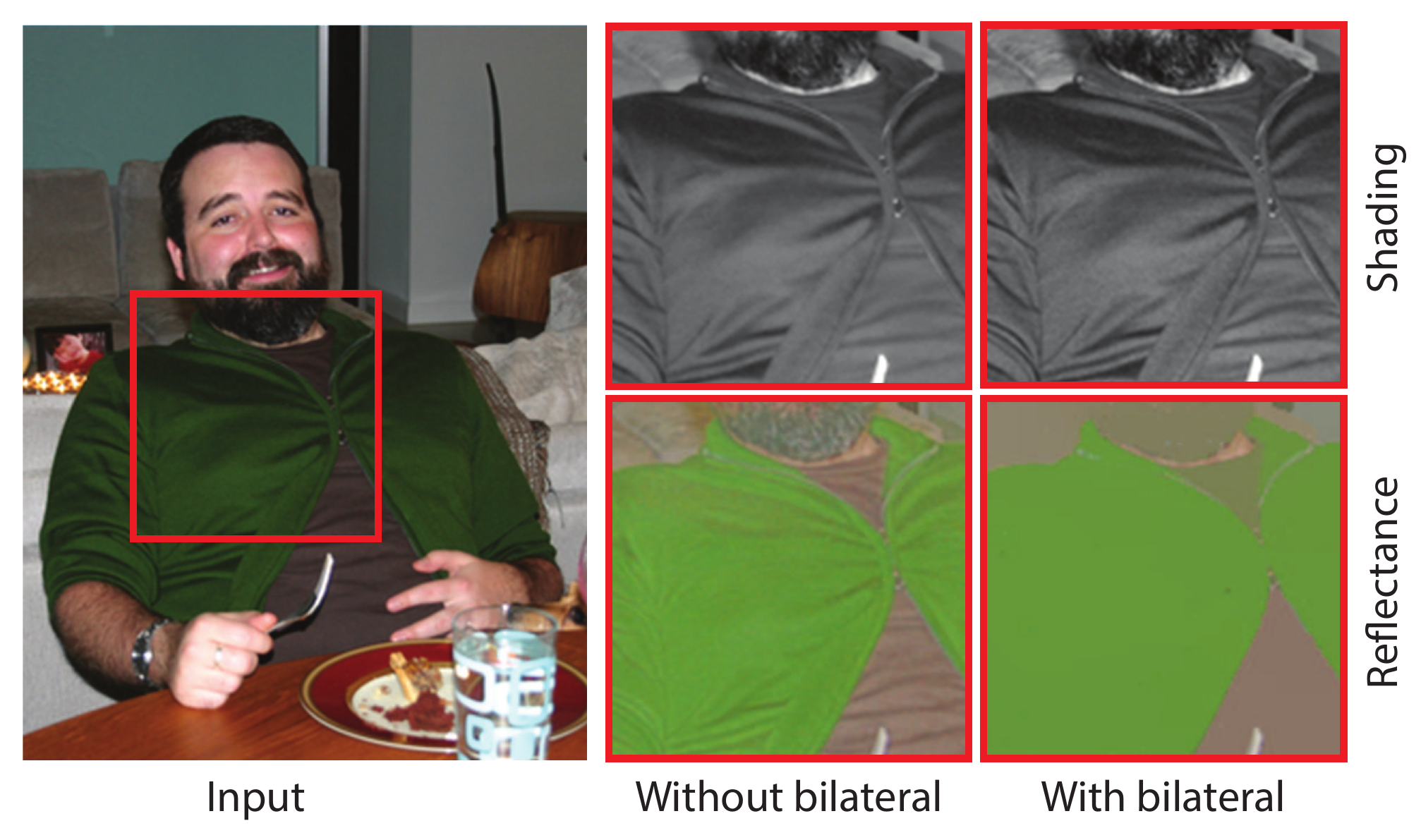}
  \caption{\label{fig:bilateral} 
  We compare the results generated using our models with and without the
  bilateral solver layer. The CNN by itself is not able to fully separate the shading
  from the reflectance and leaves the wrinkles on the shirt. Our full approach
  uses a bilateral solver to remove these low frequency shading variations from
  the reflectance image.  See numerical evaluation of the effect of bilateral
  solver layer in Table.~\ref{fig:whdr}.}
\end{figure}

As shown in Fig.~\ref{fig:bilateral} (without bilateral), the CNN by itself is
not able to fully separate the shading from reflectance. This is because the
synthetic Sintel dataset often contains images with low frequency reflectance
layers. In contrast, the reflectance of real images is usually sparse.
Therefore, training the network on this synthetic dataset encourages it to
produce reflectance images with low frequency content. Note that although we
provide a weak supervision on real images, it only constrains the network to
produce the same reflectance for a pair of real images and does not
specifically solve this problem.

We propose to address this issue by applying a bilateral solver~\cite{barron16}
on the estimated reflectance images to remove the low frequency content, while
preserving the sharp edges. 
This optimization-based approach is 
differentiable as well as efficient,  and produces results with 
better quality~\cite{barron16} than simple bilateral filtering techniques~\cite{Smith97,Tomasi98}. 

Our contribution is to integrate this
differentiable bilateral solver into our CNN as a layer (see
Fig.~\ref{fig:network}) and use it during both training and testing.
While the bilateral solver has been integrated into CNNs for applications such as depth
estimation~\cite{srinivasan18}, our approach is the first to apply it for intrinsic decomposition. 
Previous techniques such as Nestmeyer et al.~\cite{nest17} use a 
bilateral filter in post-processing. In comparison, our whole network
is trained in an end-to-end fashion.    
This integration is essential as it would
encourage the network to focus on estimating the high frequency regions in the
reflectance layer accurately and leaving the rest to the bilateral solver. Note
that, as shown in Fig.~\ref{fig:network}, we only add this bilateral solver layer to
the reflectance branch. Moreover, as discussed, the reflectance layer of the
synthetic images usually has low frequency content, and thus, filtering the
estimated reflectance on these images could potentially complicate the
training. Therefore, we only apply the bilateral solver to the estimated
reflectance for real images. Our final real loss is defined as:

\begin{equation}
  E_{\text{real}} = E_{\text{si}}(R^*_1, R^*_2) + E_{\text{si}}(R^*_1 \cdot S_2, I_2) + E_{\text{si}}(R^*_2 \cdot S_1, I_1),
  \label{eq:realFinal}
\end{equation}

\noindent where $R^*_1$ and $R^*_2$ are the result of applying bilateral solver to the estimated reflectance layers, $R_1$ and $R_2$.

In our system, the bilateral solver is formulated as follows:

\begin{equation}
R^* = \underset{\hat{R}}{\mathrm{argmin}} \ \gamma \ \sum_{i,j} W(I_i, I_j)(\hat{R}_i - \hat{R}_j)^2 + \sum_{i} (\hat{R}_i - R_i)^2
\end{equation}

\noindent where $W(I_i, I_j)$ is an exponential weight and is defined as:

\begin{equation}
W(I_i, I_j) = \exp(-||\mathbf{f}_i - \mathbf{f}_j||_2^2) , \enskip \text{where} \enskip \mathbf{f}_i = \big(\frac{x_i}{\sigma_x}, \frac{y_i}{\sigma_y}, \frac{l_i}{\sigma_l},
    \frac{u_i}{\sigma_u}, \frac{v_i}{\sigma_v}\big)
\end{equation}

Here, $I$ is the input image, $R$ is the estimated reflectance, and $\gamma$ is
a scalar that controls the smoothness of output, which we set to 12000. $W$
calculates the affinity between a pair of pixels based on the features
including spatial locations $(x, y)$ and the pixel color in LUV color space
$(l, u, v)$. $\sigma_x$, $\sigma_y$, $\sigma_l$, $\sigma_u$, and $\sigma_v$
determine the weight of each feature and we set them to 5, 5, 7, 3, and 3,
respectively. Since the output is not differentiable with respect to these
parameters, we use a numerical approach to find their values. 
Specifically, we filter our estimated reflectance
image with a set of different parameters on 100 randomly chosen scenes from the
IIW training set. We then choose the values that produce results with the
lowest WHDR score. 

\begin{figure}[t]
    \centering
    \includegraphics[width=\linewidth]{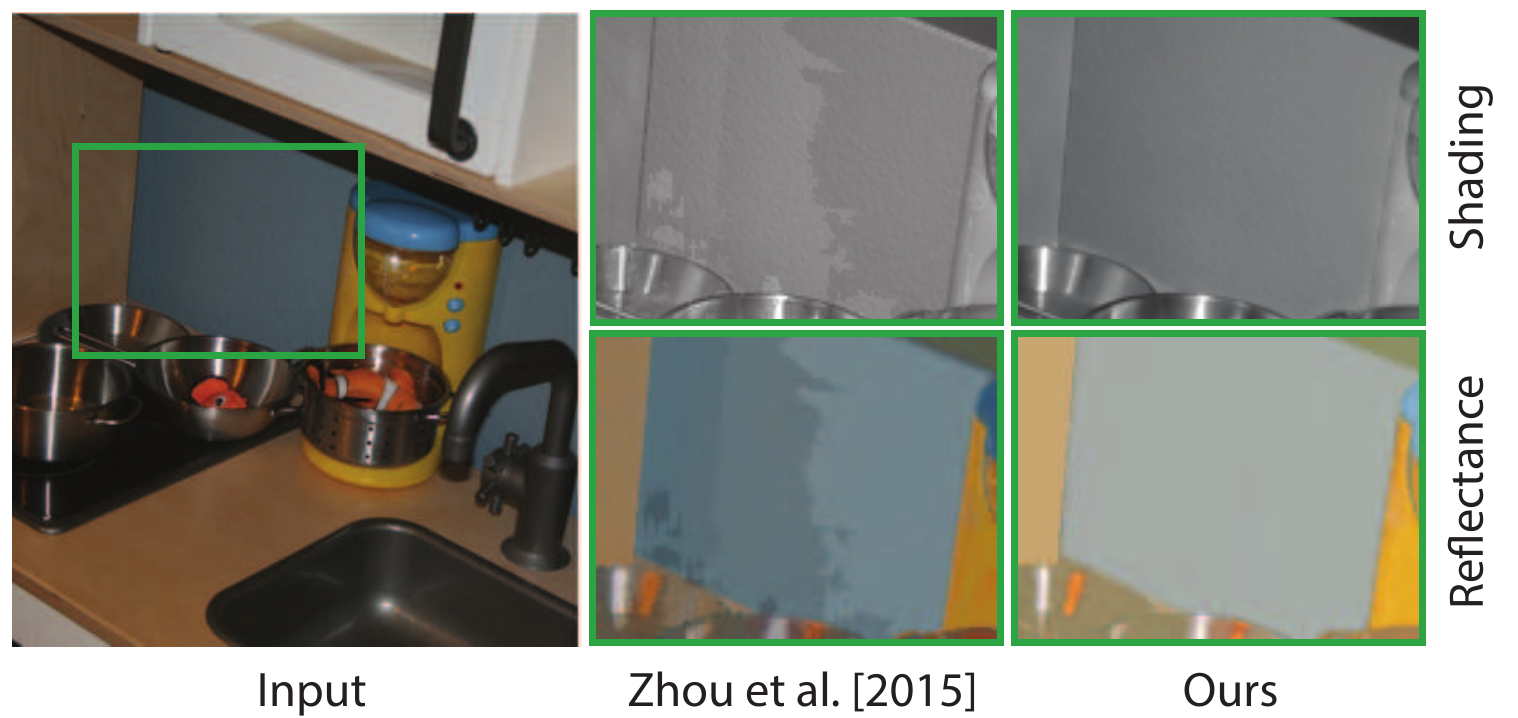}
    \caption{\label{fig:crf-comp} 
    Zhou et al.~\shortcite{zhou15} use a conditional random field (CRF) to
    cluster the reflectance pixels into a fixed number of labels.
    Therefore, their results often contain artifacts because of incorrect
    clustering, as shown in the insets. We avoid this problem by utilizing a
    bilateral solver to smooth the predicted reflectance image.   }
\end{figure}

Note that there are existing approaches that use conditional random fields
(CRF) to smooth the reflectance labellings as a post-process~\cite{bell14,
bi15} or integrate it into their learning model~\cite{zheng15}. However, these
methods perform hard clustering on a pre-defined number of labels, and thus,
produce significant artifacts in challenging cases, as shown in
Figure~\ref{fig:crf-comp}.

\begin{table}[t]
  \begin{tabular}{ l| c | c | c}
  & \makecell{Train \\ on IIW} & \makecell{Median \\ WHDR (\%)}  &  \makecell{Mean \\ WHDR (\%) } \\ \hline \hline
  Bell (2014) & \xmark & 19.63 &  20.64\\  
  Bi (2015) &  \xmark &  16.42 &  17.67\\ 
  Zhou (2015) &  \cmark & 19.13 &  19.95 \\
  Narihira (2015) & \xmark & 40.52 &  40.90  \\
  Shi (2017) &  \xmark & 54.28 &   54.44 \\
  Nestmeyer (2017) & \cmark & 16.71 & 17.69 \\
  \textbf{Ours: synthetic only}  &  \xmark &  $39.00$ & $39.18$ \\ 
  \textbf{Ours: real only}  &  \xmark &  $24.02$ & $24.74$ \\ 
  \textbf{Ours: synthetic +  real}  &  \xmark &  $20.14$ & $20.32$ \\ 
  \textbf{Ours: full model}  &  \xmark &  $\mathbf{15.86}$ & $\mathbf{17.18}$ \\ \hline \hline
  Zoran* (2015)  & \cmark  &  16.55 &   17.85 \\ 
  \textbf{Ours}* & \xmark  &  \textbf{16.01} &  \textbf{17.23} \\
  \end{tabular}
  \caption{\label{fig:whdr} Quantitative comparison on IIW dataset in terms of
  WHDR score. Note that since Zoran et al. use a different training and testing
  split, their scores are not directly comparable to other methods. Therefore,
  we compare our method against their approach separately.} 
\end{table}

\section{Results}
\label{sec:Results}
In this section, we compare our approach against several state-of-the-art
methods. Specifically, we compare against the non-learning approach of Bi et
al.~\shortcite{bi15}, the learning-based (non-DL) methods by Zhou et al.~\shortcite{zhou15}, Zoran et
al.~\shortcite{zoran15}, and Nestmeyer and Gehler~\shortcite{nest17}, and the
DL algorithms by Narihira et al.~\shortcite{narihira15} and Shi et
al.~\shortcite{shi17}. For all the approaches, we use the
source code provided by the authors. We first show the results on real-world
scenes (Sec.~\ref{sec:real-comp}) and then demonstrate the performance of our
approach on synthetic images (Sec.~\ref{sec:syn-comp}).

\begin{figure}[t]
	\centering
	\includegraphics[width=\linewidth]{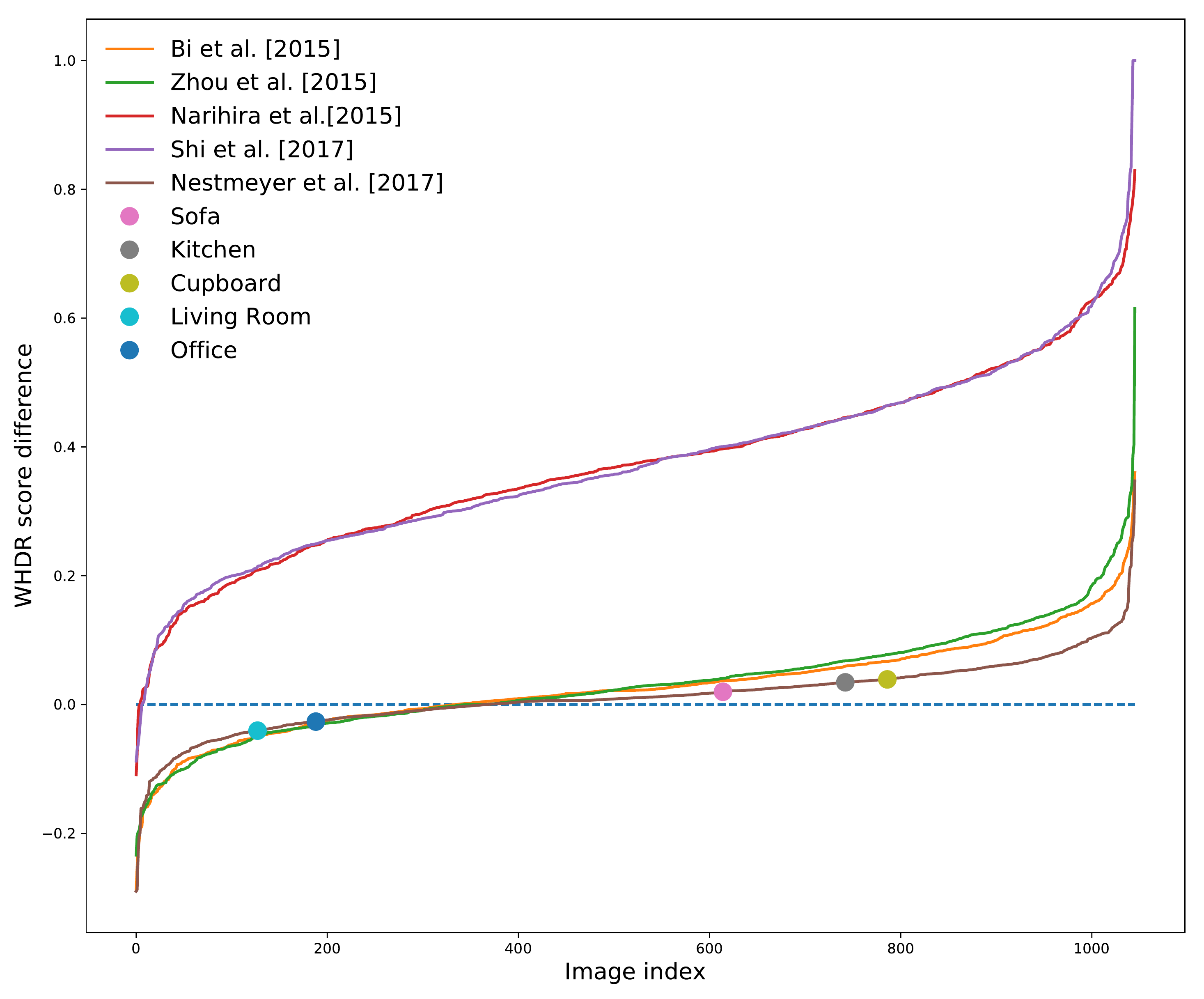}
    \caption{\label{fig:whdr-problem} For each plot, we subtract our WHDR score
    from the score of the competing approach and sort the differences for the
    1000 scenes in the IIW dataset in ascending order. Therefore, a particular
    image index corresponds to different images for each approach. Here, a
    positive difference shows that our approach is better. Our method produces
    better scores than all the other approaches in 65\% of the cases. We mark
    the images, shown in Fig.~\ref{fig:iiw-comp}, on the Nestmeyer et al.'s
    plot to demonstrate that the selected images represent the dataset.}
\end{figure}

\begin{figure*}[!ht]
	\centering
	\includegraphics[width=0.9\textwidth]{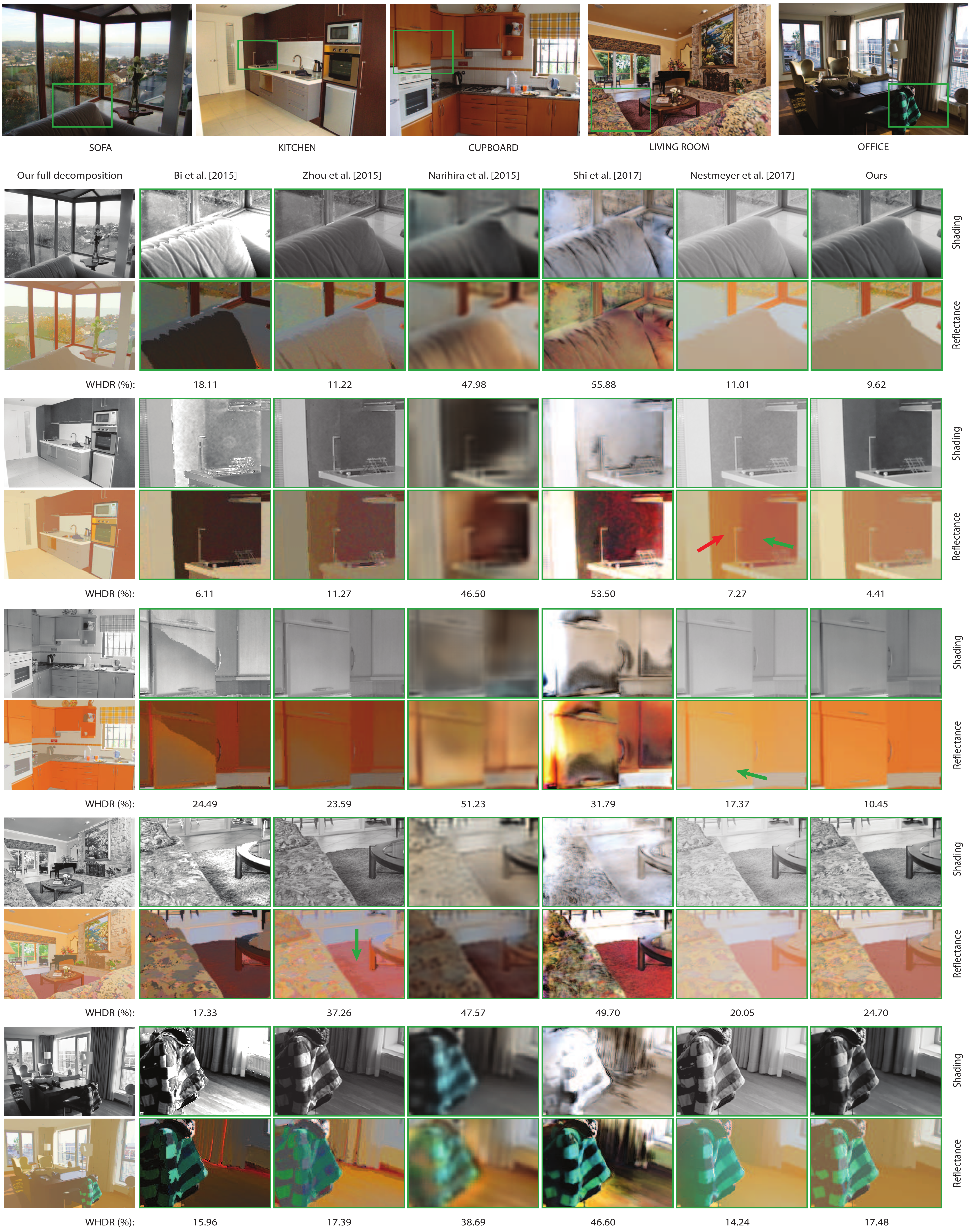}
    \caption{\label{fig:iiw-comp} Comparison against several state-of-the-art
    approaches on the IIW dataset. The WHDR score for each method is listed below
    their results. Overall, our method produces visually and numerically better
    results than the other approaches. See supplementary materials for the full
    images.}
\end{figure*}

\subsection{Comparisons on real images}
\label{sec:real-comp}

{\em Comparison on IIW Dataset --} We begin by comparing our approach
against the other methods on the IIW dataset~\cite{bell14}, which uses human
judgments of reflectance comparisons on a set of sparse pixel pairs as the
ground truth. We evaluate the quality of the estimated reflectance images for
each algorithm using the {\em weighted human disagreement rate}
(WHDR)~\cite{bell14}, as shown in Table~\ref{fig:whdr}. Although our method has
not been specifically trained on this dataset, we produce the best (lower is
better) WHDR scores compared to the other approaches. We also evaluate the effect
of each term  in Eq.~\ref{eq:loss} as well as the bilateral solver
layer. 
As seen, training using both $E_{\text{syn}}$ and $E_{\text{real}}$ 
reduces the score  
to $20.32\%$, compared to the scores of $39.18\%$ with synthetic images only  and 
$24.74\%$ with real images only.
The bilateral solver layer in our full model further decreases the 
score from $20.32\%$ to $17.18\%$.
We also plot the WHDR score differences of our and 
other approaches for all the 1046 testing images in the IIW dataset in Fig.~\ref{fig:whdr-problem}. 
As can be seen, our method produces better scores than all the other approaches 
on the majority of images.

\begin{figure}[t]
	\centering
	\includegraphics[width=\linewidth]{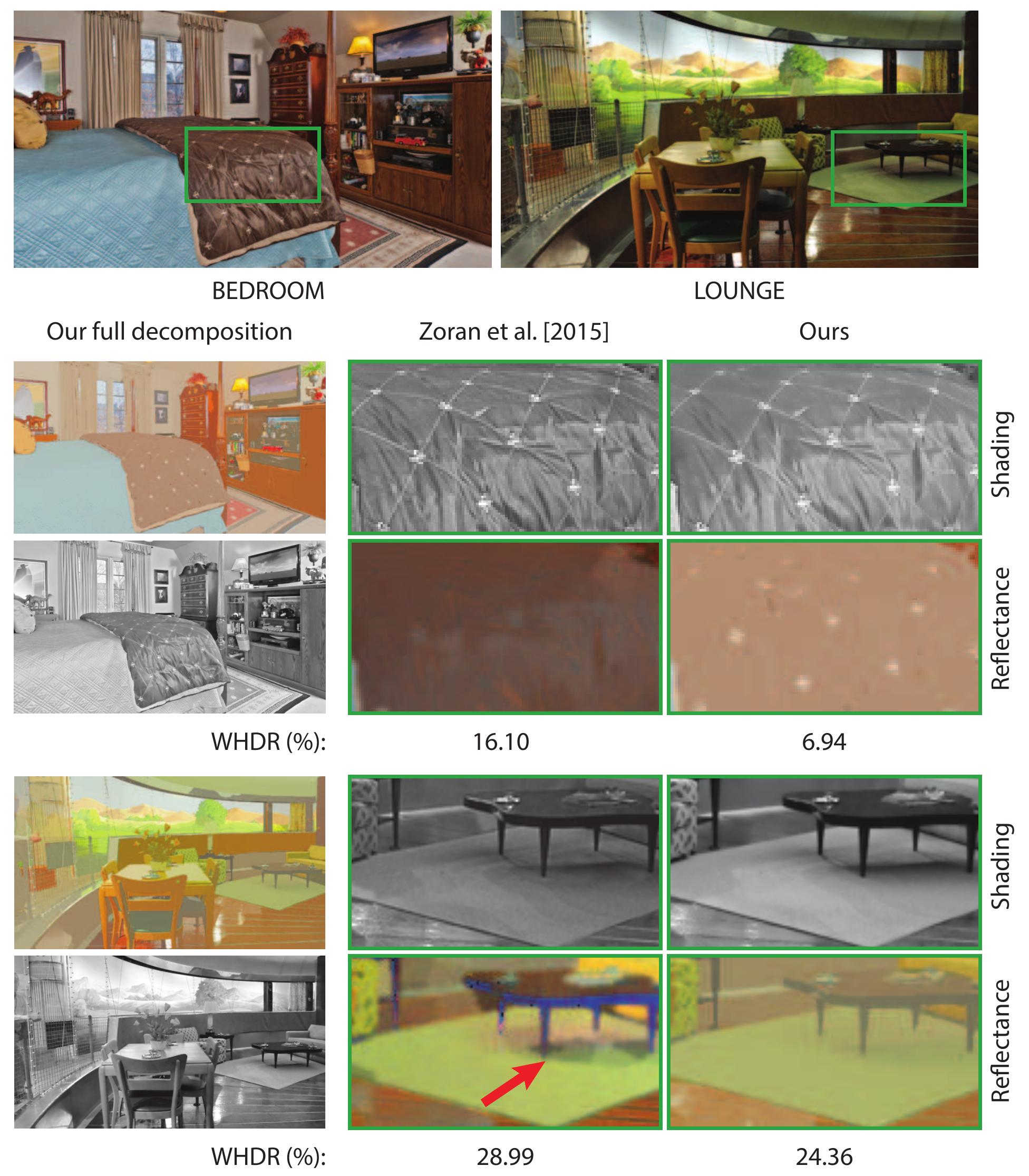}
	\caption{\label{fig:iiw-comp-zoran} Comparison against Zoran et al.'s method~\shortcite{zoran15} on two images from the IIW dataset.
	}
\end{figure}

Figure~\ref{fig:iiw-comp} compares our results against other approaches on
five representative (see Fig.~\ref{fig:whdr-problem}) scenes. Note that since Zoran
et al.'s method uses a different training and testing split, we show comparison
against their method separately in Fig.~\ref{fig:iiw-comp-zoran}. The
\textsc{Sofa} scene has complex shading variations on the surface of the sofa.
The approaches by Bi et al. and Zhou et al. use a conditional random field
(CRF) to assign a reflectance value to each pixel, and thus, their reflectance
images contain sharp color changes because of incorrect labeling. Moreover,
Zhou et al.'s method is not able to remove the wrinkles, produced by shading
variations, from the sofa. Narihira et al. and Shi et al. train their model
only on synthetic images, and thus, are not able to produce satisfactory
results on real scenes. In this case, our approach produces better results than
the other methods, but is comparable to Nestmeyer and Gehler's method.

Although the two sides of the cupboard in the \textsc{Kitchen} scene have the
same reflectance, because of large normal variation, they have different pixel
colors in the input image. While Bi et al.'s method is able to assign the
correct label to the pixels in this region, their result suffers from
inaccurate cluster boundaries, which can be clearly seen from the estimated
shading. All the other approaches are not able to produce a reflectance image
with the same color on the two sides of the cupboard. On the other hand, our
method produces a high-quality decomposition.

The \textsc{Cupboard} scene
has slight shading variations on the cabinet. None of the other approaches are
able to fully separate the shading from reflectance. However, our method
produces visually and numerically better results than other methods. 
For the \textsc{Living Room} and \textsc{Office} scene, although our WHDR
scores are slightly worse than some of the comparison methods, our method
produces visually better results by preserving better texture details in reflectance 
and recovering correct shading information.
The couch and carpet in the \textsc{Living Room} scene have complex textures and shading variations.
Bi et al., Narihira et al. and Shi et al. cannot deal with such a challenging
case and assign incorrect reflectance to the couch surface.
Zhou et al. cannot deal with shading variations on the carpet and incorrectly preserves
shading in reflectance. 
Nestmeyer et al. is able to handle the
shading variations, but it also removes the high frequency details and generates
an over-smoothed reflectance image. 
In comparison, our method is able to remove the shading
variations and preserves the texture details at the same time.
Finally, we examine the \textsc{Office} scene, which contains large shadow
areas. Despite having better WHDR scores, the estimated reflectance images by 
Bi et al. and Zhou et al. have severe artifacts in the shadow areas around the 
curtain. Narihira et al. and Shi et al. perform poorly on this challenging scene.
Moreover, Nestmeyer and Gehler's method is not able to properly handle the
shadow areas on the left side of the jacket.

\begin{figure*}[t]
  \centering
  \includegraphics[width=\textwidth]{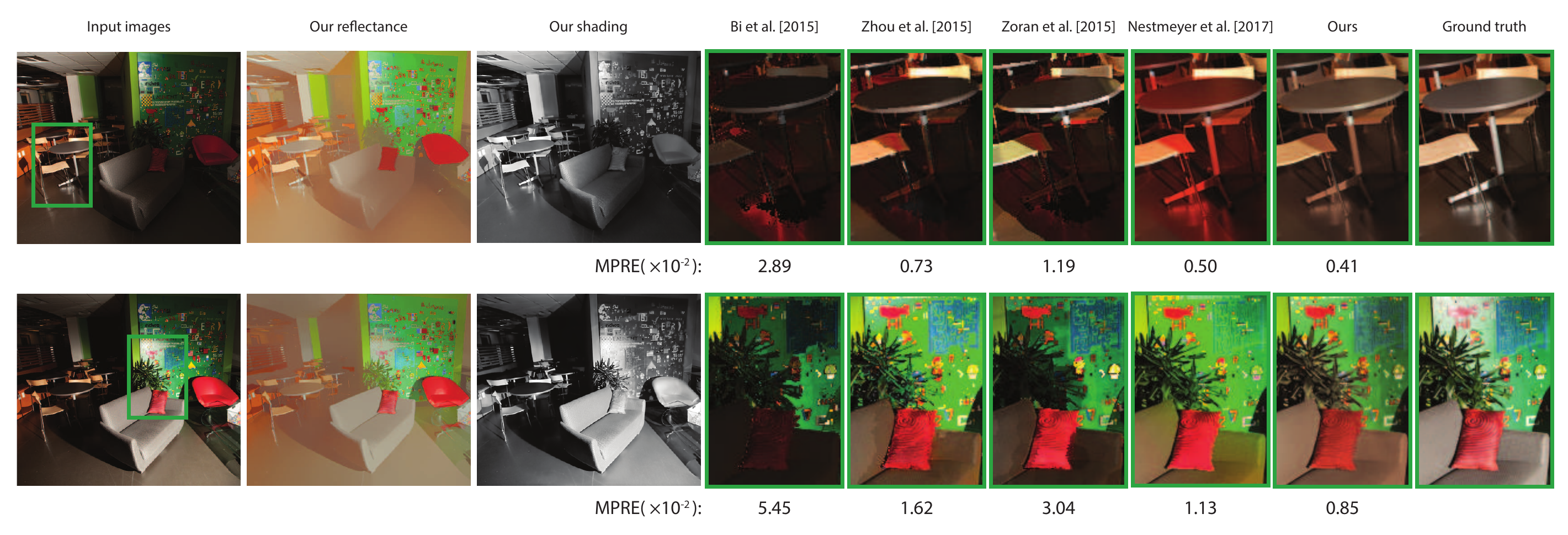}
  \caption{\label{fig:swap}  On the left, we show two images of the same scene
  taken under different illumination conditions, and our estimated reflectance
  and shading images. We show the result of swapping the reflectance images of
  the two inputs, and multiplying them with the estimated shading of each
  algorithm in the insets. Our approach produces better results than other
  methods both visually and numerically.  }
\end{figure*}


We compare our method against Zoran et al.'s algorithm in
Fig.~\ref{fig:iiw-comp-zoran}. Their method assumes that each superpixel in the
image has constant reflectance. Therefore, in the \textsc{Bedroom} scene, their
estimated reflectance lacks details on the quilt and has color variations.
Moreover, their method is not able to remove the shadows on the carpet from the
reflectance image. In comparison, our method produces better results visually
and numerically in both cases.

{\em Comparison on Illumination Invariance --} We evaluate the performance
of all the methods on generating consistent reflectance layers from images with
different illuminations. To do so, we use the dataset from Boyadzhiev et
al.~\shortcite{boyad13}, which contains four indoor scenes with each scene
consisting of a series of images captured with a tripod mounted camera from a
static scene with different illuminations. The basic idea is that the estimated
reflectance from all the images in each scene should be the same. We measure
this by using the mean pixel reconstruction error (MPRE)~\cite{zhou15}, which
is defined as follows:

\begin{equation}
    \text{MPRE}(\{I_i, R_i, S_i\}_{i=1}^{N}) = \frac{1}{N^2}\sum_{i = 1}^{N} \sum_{j=1}^{N} E_{\text{si}}(S_i\cdot R_j, I_i),
\end{equation}

\noindent where $E_{\text{si}}$ is the scale invariant MSE, which is defined in
Eq.~\ref{eq:simse}. Note that unlike Zhou et al.~\shortcite{zhou15}, here we
use the scale invariant MSE as the output of different algorithms could
potentially have different scales. We resize each input image so that its
larger dimension size is equal to $640$.

The MPRE scores for all the approaches on the four scenes are reported in
Table~\ref{table:MPRE}. As can be seen, our method produces significantly lower
MPRE scores, which demonstrates the robustness of our approach to illumination
variations. We show two images from the \textsc{Cafe} scene in
Fig.~\ref{fig:swap}. Here, we compare all the methods by swapping their
estimated reflectance images for the two scenes and multiplying them with the
estimated shading images. The result of this process should be close to the
original input images. As shown in the insets, our results 
are closer to the ground truth both visually and numerically. However, other
methods struggle to produce comparable results to the ground truth. For
example, in the top row, Nestmeyer and Gehler produce results with
discoloration. On the other hand, Bi et al., Zhou et al., and Zoran et al.
reconstruct images with incorrect shadows under the table.

\begin{table}[t]
  \centering
  \begin{tabular}{ l| c | c | c | c }
     & Cafe & Kitchen & Sofas & Uris \\ \hline \hline
   Bi (2015) &  4.47 &  4.13 &  3.48 & 3.28 \\
   Narihira (2015) & 1.50 &  1.01 & 0.87 &  1.11 \\
   Zhou (2015) & 1.68 &  0.87 &  0.49 & 0.92 \\
   Zoran (2015)  &  3.49 & 2.16 &  1.71 &  2.09\\
   Shi (2017) & 3.80 & 3.98 & 4.88 & 2.96 \\
   Nestmeyer (2017) &  2.30 &  0.63 &   0.42 &  0.66\\
   Ours &  \textbf{0.68} &  \textbf{0.57} &  \textbf{0.39} & \textbf{0.46} \\
  \end{tabular}
  \caption{\label{table:MPRE} Quantitative comparison against other methods in terms of mean pixel reconstruction error on four scenes from Boyadzhiev et al.~\shortcite{boyad13}. We factor out $10^{-2}$ from all the values for clarity.}
\end{table}

\begin{figure*}[t]
	\centering
	\includegraphics[width=\linewidth]{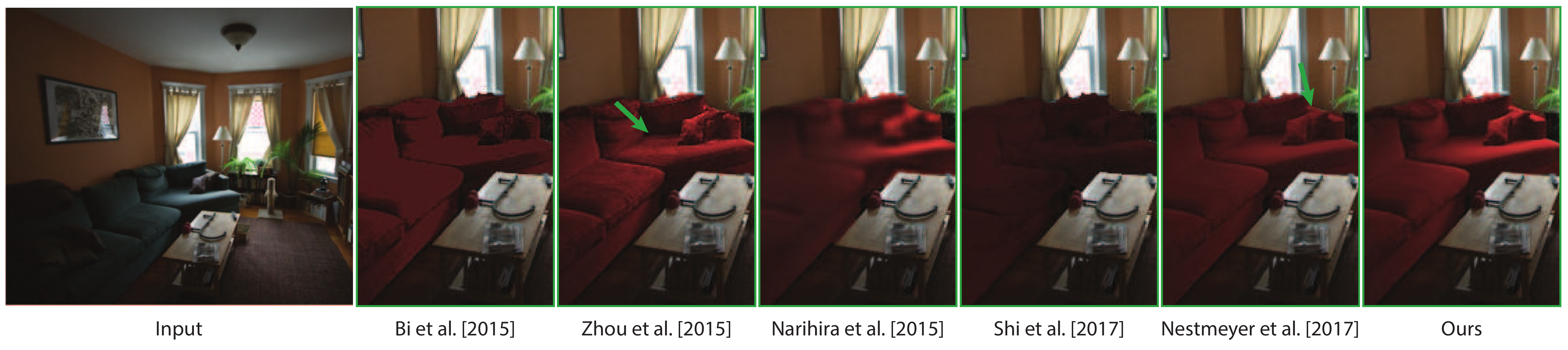}
	\caption{\label{fig:retexture} Comparisons on surface retexturing results.
    From the figure we can see that Bi et al. and Shi et al. cannot recover
    correct shading information at the retextured areas. Narihira et al.
    produce blurry results. Zhou et al. have obvious discontinuities and
    artifacts on the sofa surface, and Nestmeyer et al. miss the highlights on
    the sofa close to the window (as indicated by arrows). In comparison, our method is able to achieve
    realistic retexturing results without noticeable artifacts.
	}
\end{figure*}

\begin{figure*}[t]
  \centering
  \includegraphics[width=\textwidth]{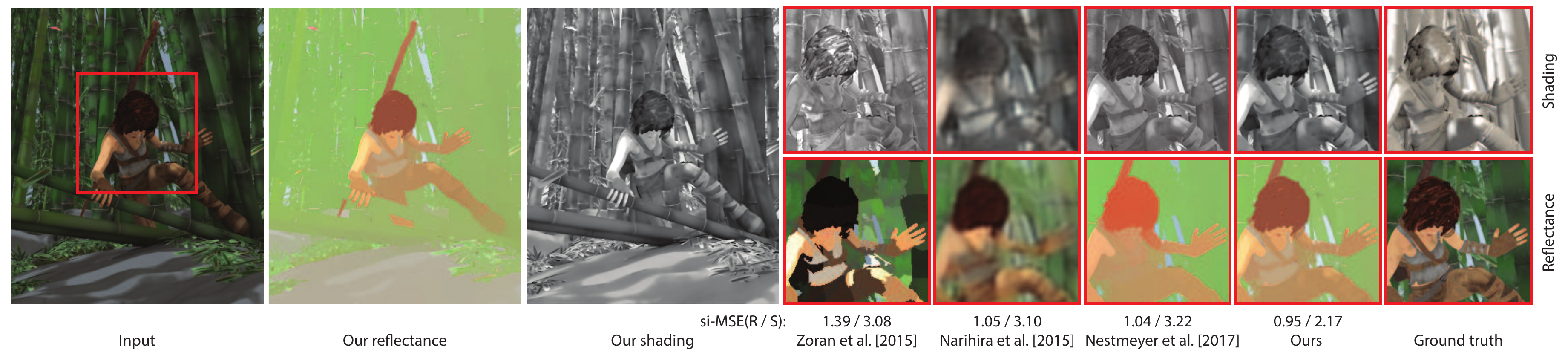}
  \caption{\label{fig:syn-visual} Comparison against the other methods on a test image from the Sintel dataset.}
\end{figure*}


{\em Comparison on Surface Retexturing --} With the shading information from 
intrinsic decomposition results, we can modify the textures of the input images
and achieve realistic image editing by multiplying the new texture by the
estimated shading. In Figure~\ref{fig:retexture}, we change the texture of 
the sofa and compare the texture editing results with the shading layer 
estimated with different approaches. From the figure, we can clearly see that 
Bi et al. and Shi et al. have incorrect lighting on the retextured areas and the
retextured images look unrealistic. Narihira et al. produces blurry results.
Zhou et al. is able to preserve the original lighting conditions, but it also
incorrectly preserves the textures in the input image, and therefore there are
obvious discontinuities and artifacts on the sofa. Nestmeyer's results miss the highlights on
the sofa close to the window. In comparison, our method
recovers the lighting information of the input image correctly and 
achieves realistic retexturing results without noticeable artifacts.

{\em Runtime Analysis --} During testing, our approach takes $1.4$ seconds to process an
input image of size $1280 \times 720$ on an Intel quad-core 3.6 GHz machine
with 16 GB memory and a GTX 1080Ti, in which $0.9$ seconds is for the
bilateral solver. Our timing is comparable to DL-methods such as Shi et al.~\cite{shi17} and Narihira et al.~\cite{narihira15},
while is significantly faster than  non-DL methods, 
which take minutes for
optimization and post-filtering.

\subsection{Comparisons on synthetic images}
\label{sec:syn-comp}

Here, we compare our approach against other methods on the test images from the
Sintel dataset. Table~\ref{fig:syn-score} shows the error for both reflectance
and shading images using the scale invariant MSE (si-MSE) and local scale
invariant MSE (si-LMSE)~\cite{grosse09}, a commonly-used metric for quality
evaluation on synthetic images. Note that the goal of our approach is producing
high-quality decompositions on real images. In contrast to the approaches by
Narihira et al. and Shi et al. that only use synthetic images for training, we
train our system on both synthetic and real images. Therefore, we did not
expect our system to perform better than these methods on synthetic images.
Nevertheless, we achieve state of the art accuracies even on this dataset,
which demonstrates that our proposed hybrid training does not significantly
reduce the accuracy on synthetic images. Compared to other approaches, our
method produces significantly better scores. We also show the results of all
the approaches on one of the Sintel images in Fig.~\ref{fig:syn-visual}. Zoran
et al.'s method produces results with noticeable discontinuities across
super-pixel boundaries. Narihira et al. is able to capture the overall
structure of reflectance, although generating blurrier results than ours.
Finally, Nestmeyer and Gehler's method is not able to reconstruct the fine
details on the cloth.

Moreover, we do another comparison on the synthetic dataset from Bonneel et al.
~\cite{Bonneel17}, and report the scores in Table~\ref{fig:bonneel}. None of
the methods in comparison train on this dataset. Different from the testing
protocol used by Bonneel et al., we use the same set of parameters for all
scenes instead of picking the best parameters for each scene.
Overall our approach produces results with higher accuracy than the other methods.


\begin{table}[t]
  \centering
  \begin{tabular}{ l| c | c | c | c |}
	    & \multicolumn{2}{c|}{si-MSE } & \multicolumn{2}{c|}{si-LMSE}  \\
	    & $R$ & $S$ & $R$ & $S$ \\ \hline \hline
   Bi (2015) & 3.00 & 3.81 & 1.48 &  1.98\\
   Narihira (2015) & \textbf{2.01} & 2.24 &  1.31 &  1.48 \\
   Zhou (2015) & 4.02 &  5.76 &  2.30 & 3.63 \\
   Zoran (2015)  & 3.67 &  2.96 &  2.17 & 1.89 \\
   Shi (2017) & 5.02 & 4.55 &  3.52 & 2.79\\
   Nestmeyer (2017) &  2.18 &  2.16 &   1.44 &  1.56\\
   Ours &  2.02 &  \textbf{1.84} & \textbf{1.30} &  \textbf{1.28} \\   
  \end{tabular}
  \caption{\label{fig:syn-score} 
    We compare all the approaches in terms of scale invariant MSE (si-MSE) and local scale invariant MSE
    (si-LMSE)~\cite{grosse09} on the test scenes from the Sintel dataset. We
    evaluate the error between the estimated and ground truth reflectance, $R$, and
    shading, $S$, images.  We factor out $10^{-2}$ from all values for clarity.}
\end{table}

\begin{table}[t]
  \centering
  \begin{tabular}{ l| c | c | c | c |}
	    & \multicolumn{2}{c|}{si-MSE } & \multicolumn{2}{c|}{si-LMSE}  \\
	    & $R$ & $S$ & $R$ & $S$ \\ \hline \hline
   Bi (2015) & 18.54 & 17.51& 3.11&  5.56  \\
   Narihira (2015) &  7.69  &  8.69 &  1.85  &  2.67  \\
   Zhou (2015) & 5.51 &  7.51 & 1.66  &  4.19 \\
   Zoran (2015)  &  14.12 & 12.07  & 3.69  &  4.82 \\
   Shi (2017) & 21.93  &  20.07 & 10.03  &  6.28\\
   Nestmeyer (2017) & \textbf{5.01}  &  7.06  &  1.57   &  3.99 \\
   Ours &   5.02 &  \textbf{6.61} &  \textbf{1.41} &  \textbf{2.43} \\   
  \end{tabular}
  \caption{\label{fig:bonneel} 
    We report  si-MSE and si-LMSE scores of all approaches on the synthetic scenes from the
    Bonneel dataset.  
    We factor out $10^{-2}$ from all values for clarity.}
\end{table}

\subsection{Limitations}
\label{ssec:Limitations}

Although our method produces better results than other approaches, in some cases 
it fails to fully separate the texture where there are large reflectance
changes from illumination. 
The \textsc{Living Room} 
scene in Figure~\ref{fig:iiw-comp} is an example of such a case where the texture appears in the estimated
shading image by our approach as well as all the other methods. Despite this,
our algorithm still produces visually better results than the competing approaches.
Moreover, in our current system we assume that the lighting is achromatic,
which is not a correct assumption in some cases. In the future, we would like
to extend our system to work with colored light sources.


%

\section{Conclusions and Future Work}
\label{sec:Conclusion}
We have presented a hybrid learning approach for intrinsic decomposition by
training a network on both synthetic and real images in an end-to-end fashion.
Specifically, we train our network on real image pairs of the same scene under
different illumination by enforcing the estimated reflectance images to be the
same. Moreover, to improve the visual coherency of our estimated reflectance
images, we propose to incorporate a bilateral solver as one of the network's
layers during both training and test stages. We demonstrate that our network is
able to produce better results than the state-of-the-art methods on a variety
of synthetic and real datasets. 

In the future, it would be interesting to apply similar ideas to more complex
imaging models such as decomposing an image into reflectance, shading, as well
as specular layers. Moreover, we would like to explore the possibility of using
the proposed hybrid training for other tasks, such as BRDF and surface normal
estimation, where obtaining ground truth on real images is difficult.

\section*{Acknowledgments}
This work was supported in part by NSF grant 1617234, ONR grant
N000141712687, a Google Research Award, and the UC San Diego Center for
Visual Computing.

\bibliographystyle{eg-alpha-doi}
\bibliography{IntrinsicEGSR}

\end{document}